\documentclass[conference]{IEEEtran}
\IEEEoverridecommandlockouts
\usepackage{cite}
\usepackage{amsmath,amssymb,amsfonts}
\usepackage{algorithmic}
\usepackage{graphicx}
\usepackage{textcomp}
\usepackage{xcolor}
\usepackage{multirow}
\def\BibTeX{{\rm B\kern-.05em{\sc i\kern-.025em b}\kern-.08em
    T\kern-.1667em\lower.7ex\hbox{E}\kern-.125emX}}

\usepackage{fancyhdr}
\renewcommand{\headrulewidth}{0pt}
\begin{document}

\title{CIAO! A Contrastive Adaptation Mechanism for Non-Universal Facial Expression Recognition\\
\thanks{This work was supported by a Starting Grant from the European Research Council (ERC) under the European Union’s Horizon 2020 research and innovation programme. G.A. No 804388, wHiSPER}
}

\author{
    \IEEEauthorblockN{Pablo Barros, Alessandra Sciutti}
    \IEEEauthorblockA{\textit{Contact Unit} \\
\textit{Italian Institute of Technology}\\
Genova, Italy \\
\{pablo.alvesdebarros,alessandra.sciutti\}@iit.it}
}



\maketitle
\thispagestyle{fancy}

\begin{abstract}
Current facial expression recognition systems demand an expensive re-training routine when deployed to different scenarios than they were trained for.
Biasing them towards learning specific facial characteristics, instead of performing typical transfer learning methods, might help these systems to maintain high performance in different tasks, but with a reduced training effort. In this paper, we propose \textbf{C}ontrastive \textbf{I}nhibitory \textbf{A}daptati\textbf{O}n (CIAO), a mechanism that adapts the last layer of facial encoders to depict specific affective characteristics on different datasets. CIAO presents an improvement in facial expression recognition performance over six different datasets with very unique affective representations, in particular when compared with state-of-the-art models. In our discussions, we make an in-depth analysis on how the learned high-level facial features are represented, and how they contribute to each individual dataset characteristics. We finalize our study by discussing how CIAO positions itself within the range of recent findings on non-universal facial expressions perception, and its impact on facial expression recognition research.
\end{abstract}

\begin{IEEEkeywords}
Facial Expression Recognition, Transfer Learning, Deep Learning
\end{IEEEkeywords}

\section{Introduction}
Automatic facial expression recognition (FER) achieved impressive results in the last decade. As in most computer vision-related fields, the advances in deep neural networks contributed to the development of super-human performances when describing facial expressions~\cite{anil2016literature}. As a consequence, several scenarios, represented by different datasets \cite{barros2018omg}, were created and investigated. Most of these scenarios are a variation of the ``in-the-wild" scene, where images are crawled from different sources of the internet, or from a closed-case environments, where facial expressions are collected based on a specific interaction context. Most of these scenarios, and subsequent recognition models, are shaped and designed based on the assumption that facial expressions are universally understood \cite{ekman1997universal}. As a consequence, in this regard, most of them use a similar labeling scheme, based on a set of given categorical concepts, or a dimensional representation of affect \cite{mollahosseini2017affectnet}. 

When these models are deployed on cross-scenarios recognition tasks, however, most of them seem to perform poorly \cite{savery2020survey} or seem to rely on high-effort transfer learning strategies to achieve good results \cite{ng2015deep}. This contradicts the idea of universal emotion perception, and yet, the typically proposed solution for this problem seems to be creating more complex models or larger datasets \cite{zhang2020emotion}. Most researchers working on FER do not properly address the issue that emotions are extremely subjective, and our understanding of affective perception is still not complete \cite{schlegel2020meta}. However, the limited generalization capabilities of the models developed on the assumption of emotion universality highlight the limits of this approximation. 

Recent findings on facial expression perception point to a non-universality of emotion understanding  \cite{jackson2019emotion}. These investigations suggest that the interpretation of affect comes from on our world understanding, which is shaped by our experiences \cite{hoemann2020expertise}. The positive results on transfer learning applied to facial expression perception \cite{xue2021transfer} could be seen as an indication that understanding facial expression as non-universal is beneficial for FER. However, most of the recent models, rely heavily on computer vision techniques that are based on the transfer of general concepts, which can hinder the development of transfer learning methods specific for FER.

 In this paper, we propose the \textbf{C}ontrastive \textbf{I}nhibitory \textbf{A}daptati\textbf{O}n (CIAO) as a novel facial expression representation adaptation method. First, we obtain a facial encoder to represent facial expressions and use them as a first approximation for the expression recognition. Then, for each small world representation, meaning the specific characteristics that will denote an emotional label in a dataset, the inhibitory layer specializes a filter, which is applied to the last convolutional layer of the encoder and optimized through a contrastive loss. As contrastive learning focuses on highlighting and identifying the differences between similar data \cite{henaff2020data}, we can quickly specify the facial representation of the encoder to the characteristics present on a dataset by the inductive bias from the labels. Our model aims to achieve an adaptation of facial representations, which improves drastically facial expression recognition while having a very low number of trainable parameters, making it ideal to be applied to underrepresented affective scenarios, where there is not enough data to learn emotional characteristics.
 
We evaluate CIAO with three experiments: first, we provide an objective investigation, based on performance measures, on the adaptability of our inhibitory layer to change the interpretation of facial expressions coming from different affective datasets. For that, we rely on an extensive experimental setup, and evaluate CIAO's performance using the following datasets: AffectNet\cite{mollahosseini2017affectnet}, FER \cite{goodfellow2013challenges}/FER+ \cite{barsoum2016training}, JAFFE \cite{lyons2020coding} and EmoReact \cite{nojavanasghari2016emoreact}. Each of these datasets has specific characteristics which make transfer learning among them very challenging. Second, we evaluate how our model impacts facial expression recognition when compared to other fine-tuning methods, and how it compares with state-of-the-art FER for each dataset. Finally, we provide a systematic feature visualization using saliency maps to highlight the impact of CIAO on the learned high-level facial features for each dataset.

Our results show that CIAO achieved competitive performance on facial FER, but most importantly, it reduces drastically the number of updatable parameters and increased the re-usability of the encoder in different scenarios. These outcomes demonstrate that our model is a novel and robust facial representation method, strongly inspired by the non-universality of affective perception. In the discussions, we analyze its impact on the application and scalability of automatic facial expression recognition, and we conclude our paper by proposing a better understanding of non-universal emotions towards affective computing.

\section{Related Work}
\label{related work}

The transfer of facial representations is currently done by many different processes. From the typical mid and high-level fine-tuning, which allows the specification of a rough representation of affect into specific ones, or using the unique characteristics from faces to leverage expression recognition \cite{liu2018automatic}. Most of these solutions, however, approach facial transfer learning as a common computer vision problem: there are a set of pixel-level structures that were learned for a certain task, and that need to be found on a novel task to decrease the training time  or reduce the need of a large number of training data \cite{yin2019feature}. Very few solutions approach this problem leveraging on a specific theory for affect understanding: emotions are subjective and their understanding varies depending on the context, not the underlying representations of faces.



\subsection{A Computer Vision View}

Transferring learned representations from a convolutional neural network seems to be extremely efficient on most computer vision tasks \cite{han2018new}. These networks showed to learn low-level features, and they specify these features in deeper layers. Adapting pre-trained networks such as the different VGG, ResNet and Inception versions towards different tasks, including facial expressions recognition \cite{shin2016deep}, became the norm. In this regard, even the adaptation of these models towards facial recognition, implemented as a Deep Face neural networks \cite{parkhi2015deep}, was proposed. Deep Face learns facial structures quite well, but when applied to facial expression recognition, the learned representations always needed to be complemented or fine-tuned in order to depict affect. Traditional fine-tuning of convolutional layers increases the cost of training these network. To change them into an alternative approach would require a change in the basic assumptions of facial expression recognition, starting from the acceptance of subjectivity of affect understanding.

\subsection{The Perspective of Affective Computing}

In an affective computing task, understanding expressions is usually not the final goal \cite{cambria2016affective}. Affect perception and affective understanding are usually separated and while the first is heavily supported by current deep neural network approaches solutions, the latter is usually a product of a psychological or physiological models of humans \cite{yin2020facial}. Disentangling perception and affect became one of the most important goals of a plethora of research \cite{mao2017learning}. Most of the proposed perception models, however, base their perception models on the powerful and successful convolutional neural network-based approaches, which results in an incomplete disentangling of learned representations. Although the recent advances on self-supervised learning, where given labels do not impact directly on the feature representation, focus on reducing the label-induced biases \cite{koujan2020real}, when applied to facial expression recognition they still learn and model specific characteristics present on the data they were trained on. Reducing these biases and the a priori assumptions would allow affective computing applications to be more adaptable, in particular when applied to dynamic areas such as social robotics \cite{faria2017towards}. Moreover, the current models are still difficult to train, adapt, and fine-tune, due to their complex training process which makes it difficult to reach a stable adaptation mechanism. 

\subsection{Why Mentioning Non-Universal Emotions?}

The idea of non-universality of emotions can seem controversial, given the repeated and established notion of facial expression recognition in today's literature \cite{shin2016deep}. However, although the current facial expression recognition models can achieve super-human performance on inferring affect from facial expressions, 
they are extremely limited to a certain dataset benchmark.
We believe it is possible to achieve a true flexible and adaptable affective understanding when we start to develop affective perception and understanding models that treat emotions as a subjective matter, that can change from scenario to scenario and from person to person. In this paper, we propose to start bridging this gap by addressing the adaptation of facial expression recognition models in different scenarios: transferring useful representations to different tasks. Treating each application scenario, and associated affective world representation, as a different task allows us to present affect as a unique experience of each of these scenarios, that share some similarities (in particular on the facial structures) and differ on their understanding. In this regard, progressive neural connections \cite{rusu2016progressive} have been applied to transfer learned knowledge over similar tasks including emotion recognition from audio \cite{gideon2017progressive}. Progressive connections, however, do not scale well when the new task has few data points and examples, which can bias the understanding of affect from previous tasks by focusing on a previously, heavily-learned affect understanding. Thus we propose a contrastive learning to provide a fast and robust adaptable mechanism for facial expression recognition.

\section{An Adaptive Inhibitory Plugin }
\label{proposed model}

To provide an adaptable mechanism for facial expression recognition we must rely on a robust facial expression representation. Up until now, there exist at least 10 years of research on convolutional-based facial representation. It was shown that convolutional layers can depict different affective characteristics on faces, particularly concentrated in the last convolutional layers. These features, however, are very much biased by the training process of these networks, and thus, rely heavily on the labeling scheme used to guide the networks' training. Our \textbf{C}ontrastive \textbf{I}nhibitory \textbf{A}daptati\textbf{O}n (CIAO) is a plugin inhibitory layer that acts as a high-level facial features mask, to be used on the last convolutional layer of a facial encoder. Our plugin layer follows the same convolutional architecture of the encoder and is trained in a supervised manner through a contrastive loss. In order to truly assess the contribution of the CIAO, we attach it to two different pre-trained encoders: a VGG-based encoder and a generative adversarial encoder. 

\subsection{A Tale of Two Encoders}

The first encoder we use is the VGGFace2 \cite{cao2018vggface2}, which is a VGG-based model trained, using supervised learning, to identify persons using a dataset with 3 million images. The inductive bias from a person identification task allows the VGGFace2 to learn specific facial features which proved to be transferable towards emotion expression recognition \cite{hussain2020real}. It consists of a VGG-based neural network with 5 convolutional blocks, followed by pooling layers. As we are interested only in its capabilities as an encoder, we use the outputs of the last convolutional layer as our encoded facial representation which is a vector of size 4608. The entire encoder has around 15 million trainable parameters.


The second encoder is the Prior-Knowledge encoder (PK) \cite{barros2019personalized}, which is a generative adversarial autoencoder that learns facial representations through a self-supervised learning scheme. The entire network has four different discriminators, each of them responsible to enforce the encoder to learn robust facial features that are distinguishable from different persons. Its original application was on the editing of facial expressions through a given arousal and valence, but the authors also demonstrated that it is a reliable facial expression encoder. The PK encoder is composed of four strided convolution layers which produce a latent representation with a size of 18432. It has a total of 4 million parameters. 



Both encoders were deliberately chosen because they were not trained with affective information. This guarantees that their learned facial representation has no inductive bias from any sort of affective label, which is the role of CIAO.

\subsection{CIAO, Contrastive Inhibitory AdapatiOn}

Our Contrastive Inhibitory AdaptatiOn (CIAO) layer aims at specifying the learned features from an encoder towards a specific scenario. As such, we propose a masked-based mechanism that can be attached to any convolutional-based encoder. CIAO takes inspiration from the inhibitory convolutional network \cite{ indolia2018conceptual} and visual attention layers \cite{rahimpour2017person}, and defines a soft attention-like mask that will deform the feature map of a convolutional channel to fit the scenarios' feature-level characteristics. Given a convolutional unit $u_{nc}^{xy}$, each masked representation $M_{nc}^{xy}$ at the position ($x$,$y$) of the $n^{th}$ receptive field in the $c^{th}$ layer is activated as:

\begin{equation}
M_{nc}^{xy} = \frac{u_{nc}^{xy}}{a_{nc} + I_{nc}^{xy}}
\end{equation}

\noindent where $I_{nc}^{xy}$ is the activation of the inhibitory connections. An integration term, $a_{nc}$, which is learned and updated together with the inhibitory connections, sets the strength of the inhibitory mask. We initialize the mask with the same values as the convolutional channel it is attached to, to guarantee a faster convergence when learning differences between the original representation and the dataset's specific representations. CIAO adds very few trainable parameters to the encoder, as it copies the last convolutional layer structure, and maintains the same dimension on the encoded representations. The updated encoded representation is then fed to a decision-making layer, that will perform the supervised classification of the facial expression. CIAO is updated together with the decision-making layer, using an adapted version of the supervised contrastive routine \cite{khosla2020supervised} defined as: 

\begin{equation}
c_{i,j}= exp(z_{i} \cdot z_{j}/t)\\
c_{i,k} = exp(z_{i} \cdot z_{k}/t)
\end{equation}

\noindent where $i$ represents each image on a training batch that shares the same labels with $j$, but has different labels than $k$, and $t$ is a temperature term defined before the training. This routine is applied for each image in a training batch, and is used to calculate the contrastive supervised loss per image:

\begin{equation}
L_{CIAO, i} = \frac{-1}{N-1}\sum_{N}^{j=1}log\frac{c_{i,j}}{c_{i,k}}
\end{equation}

\noindent where $N$ is the total number of images on that specific batch. The final loss to train CIAO can be defined as the summation of all images on that batch:

\begin{equation}
L_{CIAO} = \sum_{N}^{i=1} L_{CIAO, i}
\end{equation}

\begin{figure}
    \centering
  \includegraphics[width=0.7\columnwidth]{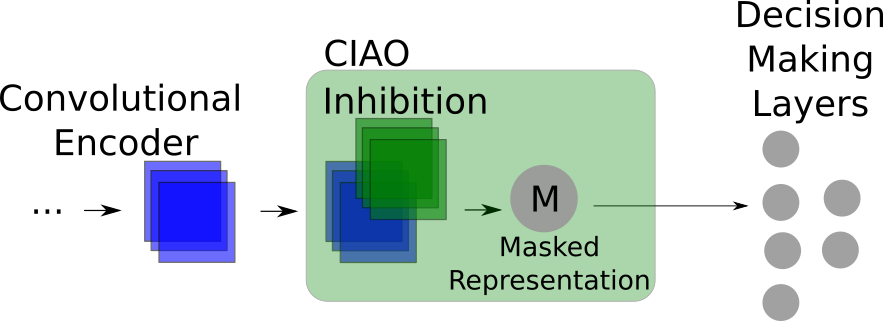}
  \caption{Conceptual architecture of CIAO: it applies an inhibitory mask onto the last convolutional channel of an encoder to generate specified facial representations update for each evaluated scenario.}
\label{fig:ciao}
\end{figure}

The final model loss can be defined as the sum of the CIAO loss and the decision-making layer loss:

\begin{equation}
L_{all} = L_{CIAO} + L_{dm}
\end{equation}

\noindent where $L_{dm}$ is the categorical cross-entropy loss when training with categorical data or the mean squared error when training with dimensional arousal and valence representation. As $L_{CIAO}$ needs a discrete way to distinguish between the sample labels, when $L_{dm}$ is the mean squared error, we discretize the continual arousal and valence values. Figure \ref{fig:ciao} illustrates the conceptual architecture of CIAO.

\section{Assessing the Impact of CIAO}
\label{experimental setup}

In order to asses the contributions of CIAO to the adaptation of facial representations, we perform two investigations. In the objective one, we assess the ability of CIAO to update the affective representations of two encoders. This is done by running an extensive experimental evaluation where we compare different transfer and fine-tuning configurations with and without CIAO for each of the evaluated encoders. Second, we use GradCAM++ to highlight the facial regions that each of the trained networks focuses on and contrast them, discussing the impact of CIAO on the adapted features.

\subsection{Affective World Representations}

For our first investigation, we rely on four datasets:



\textbf{ AffectNet~\cite{mollahosseini2017affectnet}} has more than $500$ \textit{thousand} manually annotated images, all of them crawled from the internet. Each image was annotated using two processes: first, a categorical label was given, based on one out of eight affective states (Neutral, Happiness, Sadness, Surprise, Fear, Disgust, Anger, Contempt). Also, labelers were asked to evaluate each image using continuous arousal and valence values, representing how calm/excited and negative/positive the facial expressions were, respectively. Arousal and valence were given in a range of $[-1, 1]$. Each image is presented in colors and is already pre-processed with a cropped face. 

\textbf{The FER\cite{goodfellow2013challenges} /FER+ \cite{barsoum2016training}} datasets contain the same images, with around $31,000$ gray-scaled face images crawled from the internet. The differences are in the labeling scheme. While the FER labels are given in a direct categorical manner, where each image was manually interpreted as one out of seven emotional categories (Angry, Disgust, Fear, Happy, Sad, Surprise, Neutral), the FER+ introduces a completely new labeling scheme. For each image, 10 tags were given by labelers crawled from the internet, and are presented as such. So, in FER+, each image has a label distribution, containing the sum of all the tags given to each image. 

\textbf{The JAFFE \cite{lyons2020coding}} dataset is composed of 210 images with 10 Japanese actresses performing a set of given facial expressions: Angry, Disgust, Fear, Happy, Sad, Surprise, and Neutral. Each actress performs three times each of these expressions, giving a total of 21 images per person. Each image is presented in grayscale with the face centralized. 

\textbf{The EmoReact \cite{nojavanasghari2016emoreact}} dataset has 1200 short videos collected from Youtube. The uniqueness of this dataset is that the videos contain children spontaneously reacting to other videos. Each video was annotated by several annotators base on the presence or absence of eight affective states: Curiosity, Uncertainty, Excitement, Happiness, Surprise, Disgust, Fear, and Frustration, in a way that each video has eight binary labels. We pre-processed each video by cropping the face, using the OpenCV face localizer \cite{bradski2000opencv}.

\subsection{Implementation Details and Metrics}

\begin{figure}
    \centering
  \includegraphics[width=0.8\columnwidth]{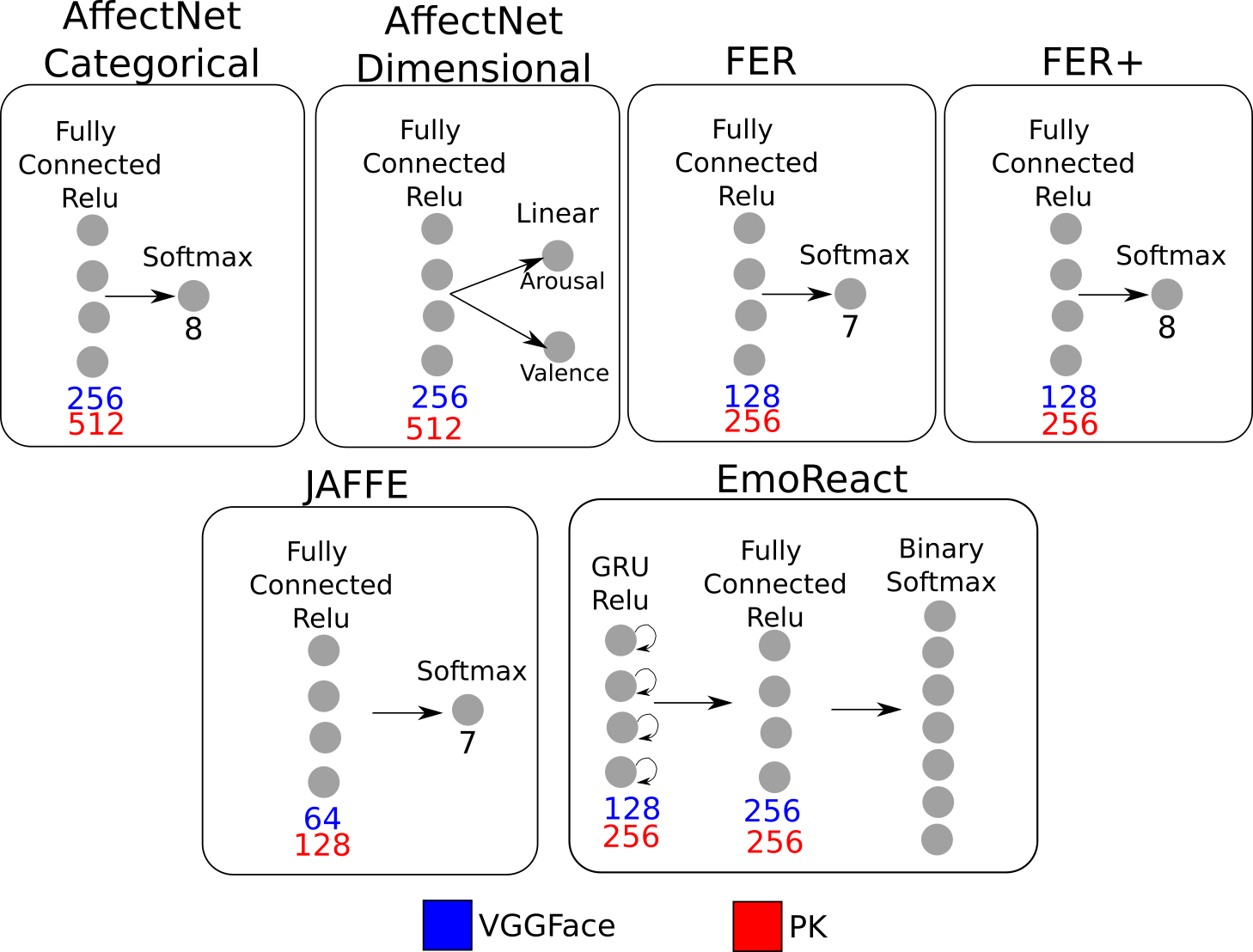}
  \caption{Final architecture of each decision-making layer for each of the datasets.}
\label{fig:decisionMaking}
\end{figure}

To evaluate the impact of CIAO in each of the datasets, we complement each encoder with a decision-making layer that was optmized to achieve the best overall result on each dataset. The final architecture, and topological and training parameters, of these networks were found using a tree-parzen search \cite{bergstra2013hyperopt}. The decision-making layers are illustrated in Figure \ref{fig:decisionMaking}.

The AffectNet, FER/FER+, and EmoReact datasets have a standard separation between training and validation samples, which we follow in all our experiments. The JAFFE evaluation follows a leave-one-emotion-out classification scheme, which is the most commonly used evaluation metric in the literature.
The categorical AffectNet, FER/FER+, and JAFFE datasets are evaluated in terms of accuracy, while the EmoReact dataset uses the F1-Score, averaged per emotional category, as a performance metric. We also make CIAO`s implementation available at our github page \footnote{https://github.com/pablovin/FaceChannel}.

\subsection{Experimental Setup}

We start our investigations with the objective experimental routine. To measure the true impact of CIAO, and the extent of its contributions on facial expression recognition performance, we run three adaptation schemes for each encoder: \textit{no re-training} (training only the decision-making layer), \textit{last conv re-training} (training the decision-making layer, and the last convolutional layer), and \textit{full re-training }(training the decision-making layer and the entire encoder). We also repeat the same training schemes, but with the presence of CIAO, totaling six trained neural networks per dataset. We run each experiment 30 times, and present the average performance for each dataset.

In our qualitative analysis, we use the GradCam++ to visualize the learned facial features that each of these networks learns, in each of the datasets. We perform a direct comparison on how the learned features vary on the two separated affective interpretations of the AffectNet and FER/FER+ datasets, and how CIAO adapts the learned features to cope with the specific characteristics of JAFFE and EmoReact.

Finally, to better understand the impact of dataset-specific adaptation we contrast the CIAO-based networks' performance with existing models for each dataset.

\section{Results}
\label{results}

\subsection{Objective Results}

The results of our objective investigation in all datasets are reported in Table  \ref{tab:FullResults}. By observing the results of the networks without CIAO, we see that the full training of both encoders is extremely beneficial and achieve the best performance in all scenarios. This is expected, as we are updating the entire model towards the specific characteristics of each dataset, over-tuning the parameters to fit the facial and affective characteristics.

When training only the last convolutional layer, the facial expression update happens only on a high-level characteristic, biased of course by the labels. In this setting, there is a general drop in performance when compared to the full re-training, which is expected, given the nature of these datasets. In particular in scenarios with very few samples, the performance drop is substantial, in particular on the PK encoder. The PK showed to be more sensitive to fine-tuning, as also displayed by the performance drop when training only the decision-making layer. In this setting, the performance is the worst of all our experiments.

When using CIAO, the general performance of all models improves on the last convolutional layer and decision-making layer settings. As CIAO forces the feature recombination to be more effective, biased by the affective information coming to form the labels, the masked attention is able to depict the facial features better than training the last convolution alone, for example. This can be seen in particular on the AffectNet valence performance, which increases to be better than the full retraining without CIAO. When using the full retraining setting, the presence of CIAO improves even further the results, showing that our contrastive optimization routine improves the general high-level representation.

\begin{table}
    \center\begin{tabular}{| c | c |c  | c | c | c|}
    \hline
        \multicolumn{6}{|c|}{\textbf{AffectNet - Dimensional (CCC)}}\\\hline
         & & \multicolumn{2}{c|}{\textbf{Arousal}} & \multicolumn{2}{|c|}{\textbf{Valence}}\\
        \hline
        \textbf{Model}& \textbf{Training} & - & \textbf{CIAO} & -   & \textbf{CIAO}    \\ \hline
        \multirow{3}{*}{VGG} & Full  &   0.46 &  0.49 &  0.50 &  0.60 \\
         & LC    &	0.35 &  0.48 &  0.40 &  0.58 \\
         & DL    &   0.29 &  0.47 &  0.31 &  0.58\\ \hline
         
         \multirow{ 3}{*}{PK} & Full  &   0.42 &  0.49 &  0.55 &  0.59 \\
         & LC    &	0.30 &  0.47 &  0.38 &  0.55 \\
         & DL    &   0.20 &  0.45 &  0.15 &  0.54\\
            \hline
            
        \multicolumn{6}{|c|}{\textbf{AffectNet - Categorical (Accuracy)}}\\\hline
        
        \textbf{Model}& \textbf{Training} & \multicolumn{2}{|c|}{\textbf{-}} & \multicolumn{2}{|c|}{\textbf{CIAO}}      \\ \hline
        
        \multirow{ 3}{*}{VGG} & Full  &   \multicolumn{2}{c|}{60.1} & \multicolumn{2}{c|}{65.7}  \\
        
         & LC &	 
         \multicolumn{2}{c|}{50.3} & \multicolumn{2}{c|}{63.4} \\
         
         & DL &	 
         \multicolumn{2}{c|}{35.2} & \multicolumn{2}{c|}{62.9} \\
            \hline    
            
        \multirow{ 3}{*}{PK} & Full  &   \multicolumn{2}{c|}{59.9} & \multicolumn{2}{c|}{63.6}  \\

         & LC &	 
         \multicolumn{2}{c|}{45.3} & \multicolumn{2}{c|}{60.2} \\
         
         & DL &	 
         \multicolumn{2}{c|}{15.2} & \multicolumn{2}{c|}{59.9} \\
            \hline    
            
        \multicolumn{6}{|c|}{\textbf{FER (Accuracy)}}\\\hline
        
        \textbf{Model}& \textbf{Training} & \multicolumn{2}{|c|}{\textbf{-}} & \multicolumn{2}{|c|}{\textbf{CIAO}}      \\ \hline
        
        \multirow{ 3}{*}{VGG} & Full  &   \multicolumn{2}{c|}{72.5} & \multicolumn{2}{c|}{76.1}  \\
        
         & LC &	 
         \multicolumn{2}{c|}{63.2} & \multicolumn{2}{c|}{69.3} \\
         
         & DL &	 
         \multicolumn{2}{c|}{55.3} & \multicolumn{2}{c|}{68.2} \\
            \hline    
            
        \multirow{ 3}{*}{PK} & Full  &   \multicolumn{2}{c|}{70.1} & \multicolumn{2}{c|}{75.4}  \\

         & LC &	 
         \multicolumn{2}{c|}{59.2} & \multicolumn{2}{c|}{70.1} \\
         
         & DL &	 
         \multicolumn{2}{c|}{28.2} & \multicolumn{2}{c|}{70.2} \\
            \hline    
            
              \multicolumn{6}{|c|}{\textbf{FER+ (Accuracy)}}\\\hline
        
        \textbf{Model}& \textbf{Training} & \multicolumn{2}{|c|}{\textbf{-}} & \multicolumn{2}{|c|}{\textbf{CIAO}}      \\ \hline
        
        \multirow{ 3}{*}{VGG} & Full  &   \multicolumn{2}{c|}{91.7} & \multicolumn{2}{c|}{94.5}  \\
        
         & LC &	 
         \multicolumn{2}{c|}{83.5} & \multicolumn{2}{c|}{90.1} \\
         
         & DL &	 
         \multicolumn{2}{c|}{72.3} & \multicolumn{2}{c|}{89.3} \\
            \hline    
            
        \multirow{ 3}{*}{PK} & Full  &   \multicolumn{2}{c|}{88.9} & \multicolumn{2}{c|}{93.2}  \\

         & LC &	 
         \multicolumn{2}{c|}{81.1} & \multicolumn{2}{c|}{89.4} \\
         
         & DL &	 
         \multicolumn{2}{c|}{69.3} & \multicolumn{2}{c|}{90.1} \\
            \hline

              \multicolumn{6}{|c|}{\textbf{JAFFE (Accuracy)} }\\\hline
        
        \textbf{Model}& \textbf{Training} & \multicolumn{2}{|c|}{\textbf{-}} & \multicolumn{2}{|c|}{\textbf{CIAO}}      \\ \hline
        
        \multirow{ 3}{*}{VGG} & Full  &   \multicolumn{2}{c|}{92.1} & \multicolumn{2}{c|}{93.4}  \\
        
         & LC &	 
         \multicolumn{2}{c|}{73.4} & \multicolumn{2}{c|}{92.1} \\
         
         & DL &	 
         \multicolumn{2}{c|}{69.3} & \multicolumn{2}{c|}{91.5} \\
            \hline    
            
        \multirow{ 3}{*}{PK} & Full  &   \multicolumn{2}{c|}{83.3} & \multicolumn{2}{c|}{85.2}  \\

         & LC &	 
         \multicolumn{2}{c|}{92.1} & \multicolumn{2}{c|}{80.3} \\
         
         & DL &	 
         \multicolumn{2}{c|}{27.8} & \multicolumn{2}{c|}{79.8} \\
            \hline           
            
                  \multicolumn{6}{|c|}{\textbf{EmoReact (F1-Score)}}\\\hline
        
        \textbf{Model}& \textbf{Training} & \multicolumn{2}{|c|}{\textbf{-}} & \multicolumn{2}{|c|}{\textbf{CIAO}}      \\ \hline
        
        \multirow{ 3}{*}{VGG} & Full  &   \multicolumn{2}{c|}{0.7} & \multicolumn{2}{c|}{0.73}  \\
        
         & LC &	 
         \multicolumn{2}{c|}{0.6} & \multicolumn{2}{c|}{0.7} \\
         
         & DL &	 
         \multicolumn{2}{c|}{0.58} & \multicolumn{2}{c|}{0.69} \\
            \hline    
            
        \multirow{ 3}{*}{PK} & Full  &   \multicolumn{2}{c|}{0.74} & \multicolumn{2}{c|}{0.76}  \\

         & LC &	 
         \multicolumn{2}{c|}{0.63} & \multicolumn{2}{c|}{0.72} \\
         
         & DL &	 
         \multicolumn{2}{c|}{0.26} & \multicolumn{2}{c|}{0.69}  \\  \hline

\label{tab:FullResults}
    \end{tabular} 
    \caption{Results of our experiments in all of the six datasets, with and without CIAO, on the three configurations: Full training (Full), Last Convolutional Layer (LC), and Decision Layer (DL).}

\end{table}

\subsection{Qualitative Analysis}

\begin{figure}
    \centering
  \includegraphics[width=0.65\columnwidth]{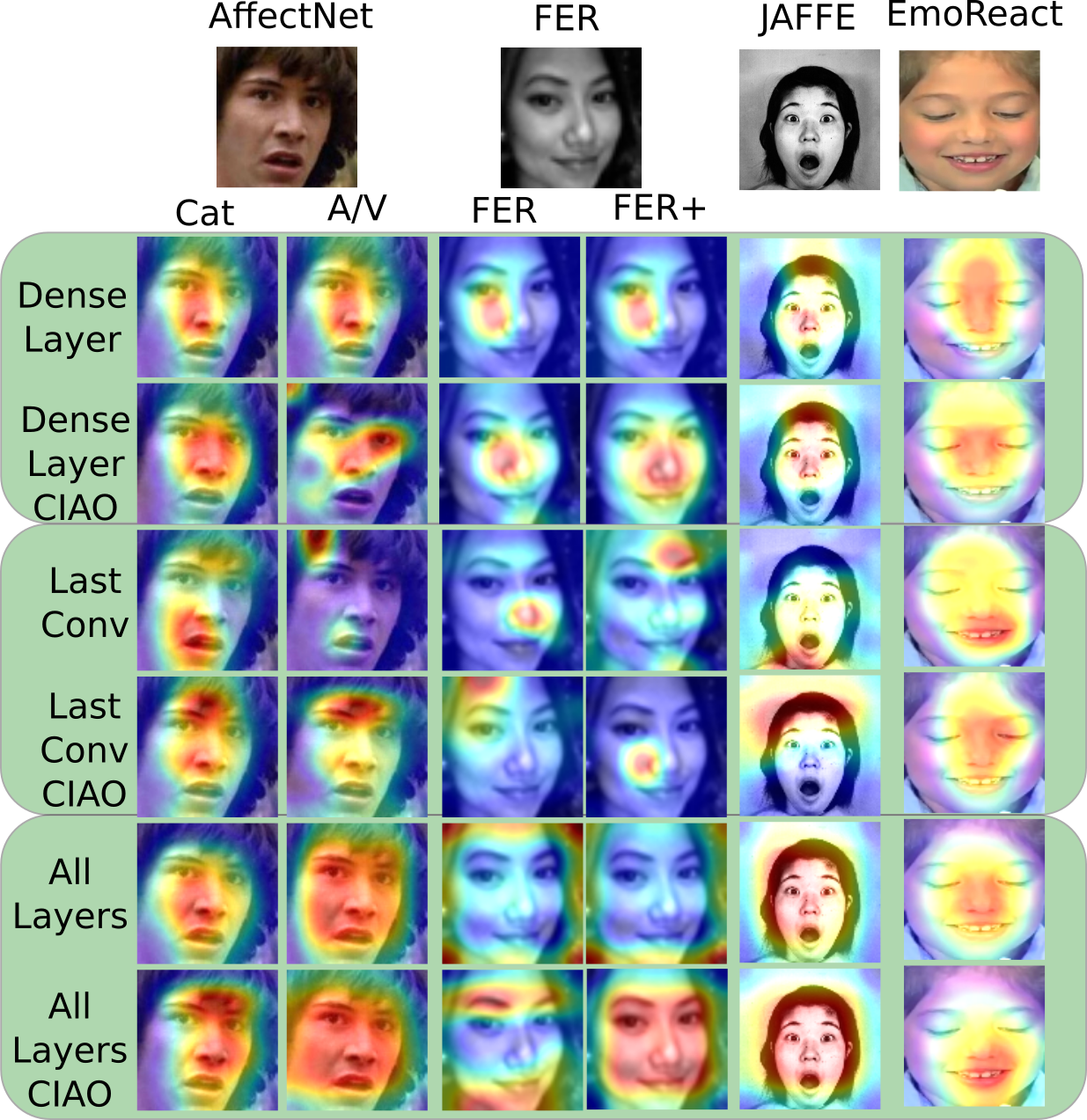}
  \caption{GradCam++ visualization of the last convolutional layer of all the networks trained with the VGG encoder.}
\label{fig:vis1}
\end{figure}

The performance improvement gain of CIAO maintains a constant rate between the two world interpretations of AffectNet (Categorical and Arousal/Valence), and FER/FER+ datasets, indicating that CIAO is able to apply the inductive bias of the specific affective interpretation much better than the traditional fine-tuning schemes. When re-training the entire network, the encoders with CIAO achieve the best results, but only with a slight improvement when compared to the non-CIAO encoders. When re-training the entire network, the feature representation is directly affected by the entire inductive bias present on the dataset, which probably directs the encoder to a similar representation obtained when CIAO is not present. In general, when re-training the network, the encoder will lose its generalization capability without a full re-training. When observing the convolutional features of both encoders, illustrated in Figure \ref{fig:vis1} for the VGG, we note the impact of the different adaptation mechanisms on the encoders. In all cases, the area of the image which highlights the last convolutional filter increases as we train more and more layers. Also, we observe the direct impact of CIAO on these features: CIAO specializes them into parts of the face that are more relevant for that specific dataset, which relates directly to the improvement on performance observed when CIAO is present. This is even  clearer when comparing the AffectNet and FER/FER+ different labeling schemes. When CIAO is present, we see the inductive biases of the different labels acting differently on the feature selection. The same effect can also be observed on the specific characteristics of the JAFFE and EmoReact datasets.

\subsection{State-of-the-art comparison}

\begin{table}[h]
    \center\begin{tabular}{ c c c c c}
    \hline
        \multicolumn{5}{c}{AffectNet}\\\hline
        \multicolumn{3}{c}{A/V} & \multicolumn{2}{c}{Categorical}\\ \hline
        Model & A & V & Model & Acc      \\ \hline
        AlexNet	&0.34 &	0.6 & PSR & 60.68\\
        MobileNet&	0.48&	0.57& RAN & 59.5\\
        VGG+GAN	&	\textbf{0.54}&	\textbf{0.62}& ESR-9& 59.3\\
        $VF_{CIAO}$	&0.50&0.60&$VF_{CIAO}$&\textbf{65.7}\\
        $VDL_{CIAO}$	& 0.47 & 0.58 &$VDL_{CIAO}$&62.9\\
        
            \hline
        \multicolumn{5}{c}{FER}\\\hline

        \multicolumn{3}{c}{FER} & \multicolumn{2}{c}{FER+}\\ \hline
        Model & \multicolumn{2}{c}{Acc} & Model & Acc      \\ \hline
         ResMaskNet & \multicolumn{2}{c}{\textbf{76.8}} & PSR & 89.75 \\
        Inception & \multicolumn{2}{c}{ 72.4} & RAN & 89.16 \\
        DeepEm. & \multicolumn{2}{c}{70.0} &SENet& 88.8 \\ 
        $VF_{CIAO}$ &\multicolumn{2}{c}{ 76.1} & $VF_{CIAO}$ &\textbf{94.5} \\
        $PKDL_{CIAO}$ & \multicolumn{2}{c}{70.2} & $PKDL_{CIAO}$ & 90.1 \\ 
        \hline

        \multicolumn{3}{c}{JAFFE} & \multicolumn{2}{c}{EmoReact}\\\hline
        Model & \multicolumn{2}{c}{Acc} & Model & F1-Score      \\ \hline
         DeepEm. & \multicolumn{2}{c}{92.2} & RBFSVM & 0.69 \\
        SalientPatch & \multicolumn{2}{c}{91.8} & SVM & 0.66 \\
         $VF_{CIAO}$ &\multicolumn{2}{c}{ \textbf{93.4}} & $PK_{CIAO}$& \textbf{0.76} \\
        $VDL_{CIAO}$ & \multicolumn{2}{c}{91.5} & $PKDL_{CIAO}$ & 0.69 \\ 
         \hline

    \end{tabular} 
    \caption{Performance comparison between the best models with CIAO (VGG full retraining - VF; VGG decision layer - VDL; PK decision layer - PKDL)  and the state-of-the-art models for each of the evaluated datasets: AffectNet in terms of Arousal(A) and Valence (V) CCC, and categorical accuracy (acc); FER/FER+ and JAFFE in terms of categorical accuracy (acc); and EmoReact in terms of F1-Score.}

\label{tab:results}
\end{table}

CIAO did improve the general performance on both encoders, and when re-training the entire network, achieved the best results in our experiments. When compared with state-of-the-art models for each dataset, reported in Table \ref{tab:results}, the full-retraining with CIAO shows to be very competitive. Yet, when training only the decision layer, which yields less parameter updates, the encoders with CIAO also provide a very competitive result, achieving better performance than some of the state-of-the-art models. When compared with the models evaluated on the AffectNet arousal and valence label set, the VGG full-training + CIAO ($VF_{CIAO}$) achieve the second-best results, just behind Kollias et al. \cite{kollias2020deep}, which employs a VGG encoder re-trained by a generative adversarial network. Therefore, it increases drastically the computational cost and training effort. When compared to the fine-tuning of the AlexNet and MobileNet \cite{hewitt2018cnn}, $VF_{CIAO}$ achieves better results. Comparing with the categorical set of the AffectNet, $VF_{CIAO}$ achieves the best result. In particular, when compared with the PSR \cite{vo2020pyramid} and RAN \cite{wang2020region}, which apply different attention-based mechanisms to specialize the learned features, our model presents a more effective solution. Also in this setting, the ESR-9 \cite{siqueira2020efficient} network presents an ensemble of convolutional neural networks that is expensive to train and presented a lower accuracy.

On the FER and FER+ datasets, the $V_{CIAO}$ also presents very competitive results. On the FER, it presents the second-best accuracy to date, only losing to the ensemble of masked networks \cite{goodfellow2013challenges}, which is extremely costly to train. When compared to traditional fine-tuning of deep neural networks, such as the Inception \cite{pramerdorfer2016facial} and DeepEmotion \cite{minaee2019deep}, the  $V_{CIAO}$ presents the best results. On the FER+ dataset,  $V_{CIAO}$ has the best results to date, compared with attention-based networks PSR \cite{vo2020pyramid} and RAN\cite{wang2020region} networks, and a cross-modal transfer learning obtained by the SENet \cite{albanie2018emotion} network.

On the JAFFE and EmoReact datasets, the full retraining with CIAO of the VGG $V_{CIAO}$ and the PK $PK_{CIAO}$ obtained the best results. While in terms of performance, the $V_{CIAO}$  achieves a better result than the fine-tuning of the DeepEmotion \cite{minaee2019deep} and the attention-based salient patch neural network \cite{happy2014automatic} on JAFFE, $PK_{CIAO}$ presents a big improvement over the use of different SVM-based approaches \cite{nojavanasghari2016emoreact} on the EmoReact dataset. 

It is important to notice that while we compare with the best networks on our experiments, the final results of applying CIAO to the dense layer only training scheme are not much worse. The latter approach reduces drastically the number of training parameters and makes the encoder fully re-usable.
So the trade-off between performance and adaptability is not compromised when CIAO is present.

\section{Discussions}
\label{discussions}

Our results demonstrate the positive impact of CIAO on the performance of the two state-of-the-art encoders. It is important to stress that the two encoders were not trained with affective information, so the performance boost is directly related to CIAO adding the inductive bias of each dataset into the encoded representation. To better understand the impact of dataset-specific adaptation, we discuss the effect of CIAO on the performance and the training effort for each of the scenarios. Also, we elaborate on how CIAO achieves flexibility on facial expression recognition, that is in line with the non-universal emotions concept, and why this is very important for the future deployment of affective computing in real-world scenarios.

 \subsection{The impact of CIAO}


When training the last convolutional layer, there is a large boost on performance, for both PK and VGGFace, which means that the main adaptation of the facial encoders happens there \cite{zhao2019visualizing}. When observing the results of the networks with CIAO, we see a clear performance improvement in all scenarios. Moreover, with CIAO, the differences in performance among all the settings are reduced drastically. This shows that the adaptation that CIAO proposes is an efficient method to be used instead of the full adaptation.


\subsection{The real cost of adaptation}

\begin{figure}
    \centering
  \includegraphics[width=0.7\columnwidth]{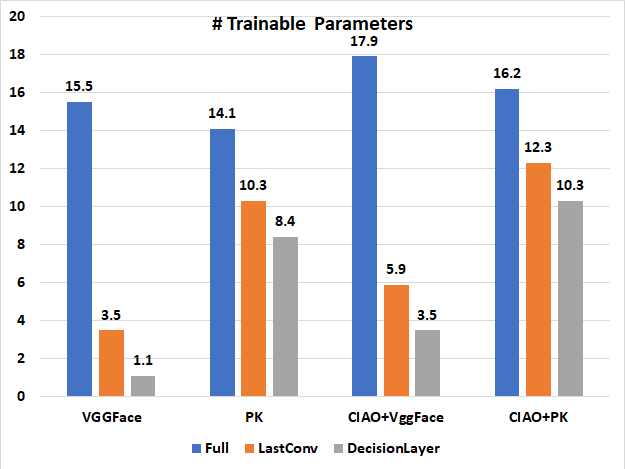}
  \caption{Averaged number of parameters, for all scenarios, per network type in millions.}
\label{fig:parameters}
\end{figure}

When analyzing the number of updated parameters per each network, averaged over all the dataset variations per network in Figure \ref{fig:parameters}, we have an in-depth understanding of the impact of CIAO on facial expression adaptation. Re-training all layers of all networks achieve the best results, but the amount of trainable parameters is much higher. Re-training the last-convolutional layer of the encoders achieve a similar performance when training only the decision layer and CIAO, but with the disadvantage that the encoder representation changes, and thus, becomes specialized towards the evaluated dataset. Transfer learning, in this case, is damaging the re-usability of the encoder in different scenarios, which is not the case when we only train the masked representations of CIAO.

\subsection{A Non-Universal Adaptation Mechanism}

The presence of CIAO allows the encoders to achieve much higher performance in all the tasks, especially when the encoder is not affected by the training scheme. This means the encoder still carries the original facial representation it originally has, but CIAO adds a biasing layer towards that specific affective world that improves the encoder capability to cope with the specific characteristics of that dataset. In general, this is an extremely important feature for a flexible facial expression recognition system, in particular when used in real-world scenarios. Decoupling the affective understanding per dataset  allows CIAO to adapt a non-universal understanding of emotions, and allows our model to be used on constrained and limited hardware, such as social robots. Also, by reducing the inductive bias from the encoder, allows it to be re-used in different tasks without major changes, facilitating its use in other scenarios. 

\subsection{Future Work}

Our entire evaluation scheme was based on offline learning of facial representations. We will work towards adapting CIAO to be suitable for online learning, to make it ideal for restricted scenarios, such as social robotics and mobile applications. Also, the need for supervised learning is important towards biasing the understanding of each affective characteristic present in a scenario, but combining strong supervision with a self-supervised training routine could help on reducing the learning effort when CIAO is applied.

\section{Conclusion}
\label{conclusions}

In this paper, we presented a novel framework called CIAO, a Contrastive Adaptive Inhibitory layer to adapt facial representations towards unique affective worlds. Our model leverages on facial encoders and learns by using a contrastive loss to adapt high-level facial features towards specific characteristics in different datasets. Our experiments involve the training and evaluation of CIAO when applied to two facial encoders, and we observe an improvement in performance in all our evaluated tasks. We assess the capability of CIAO to adapt towards different interpretations of the same data, using the AffectNet and the FER/FER+ datasets, and towards specific characteristics of the JAFFE and EmoReact datasets. Beside improving FER performance when compared to traditional fine-tuning schemes, the presence of CIAO made the facial expression recognition on both encoders competitive, and in most cases even better, when compared with the state-of-the-art models on each of these datasets. We also discussed how CIAO approaches the concept of non-universal affect by providing a flexible solution for multi-scenario applications.

\bibliography{references}
\bibliographystyle{IEEEtran}

\end{document}